\documentclass{ecai} 
\usepackage{latexsym}
\usepackage{amssymb}
\usepackage{amsmath}
\usepackage{algorithm}
\usepackage{algorithmic}
\usepackage{amsthm}
\usepackage{blindtext}
\usepackage{enumitem}
\usepackage{xcolor}
\usepackage{graphicx}
\usepackage{float}
\usepackage[hidelinks]{hyperref}

\usepackage{tabularx}
\usepackage{caption}
\usepackage{booktabs}
\usepackage{multirow}
\usepackage{subfig}

\usepackage{rotating}
\usepackage{longtable}

\newcommand{\BibTeX}{B\kern-.05em{\sc i\kern-.025em b}\kern-.08em\TeX}
\usepackage[skip=3pt]{caption}

\begin{document}

%%%%%%%%%%%%%%%%%%%%%%%%%%%%%%%%%%%%%%%%%%%%%%%%%%%%%%%%%%%%%%%%%%%%%%%%

\begin{frontmatter}

%%% Use this command to specify your submission number.
%%% In doubleblind mode, it will be printed on the first page.

\paperid{123} 

%%% Use this command to specify the title of your paper.

\title{{\sc \textbf{TopFormer}}: Topology-Aware Authorship Attribution 
of Deepfake Texts with Diverse Writing Styles}

%%% Use this combinations of commands to specify all authors of your 
%%% paper. Use \fnms{} and \snm{} to indicate everyone's first names 
%%% and surname. This will help the publisher with indexing the 
%%% proceedings. Please use a reasonable approximation in case your 
%%% name does not neatly split into "first names" and "surname".
%%% Specifying your ORCID digital identifier is optional. 
%%% Use the \thanks{} command to indicate one or more corresponding 
%%% authors and their email address(es). If so desired, you can specify
%%% author contributions using the \footnote{} command.

\author[A]{\fnms{Adaku}~\snm{Uchendu}\thanks{Corresponding Author. Email: adaku.uchendu@ll.mit.edu.}}
\author[B]{\fnms{Thai}~\snm{Le}}
\author[C]{\fnms{Dongwon}~\snm{Lee}} 

\address[A]{MIT Lincoln Laboratory, Lexington, MA, USA}
\address[B]{Indiana University, Bloomington, IN, USA}
\address[C]{The Pennsylvania State University, University Park, PA, USA }

%%% Use this environment to include an abstract of your paper.

\begin{abstract}
Recent advances in Large Language Models (LLMs) have enabled the generation of open-ended high-quality texts, that are non-trivial to distinguish from human-written texts. We refer to such LLM-generated texts as  \emph{deepfake texts}.
There are currently over 72K text generation models in the huggingface model repo. As such, users with malicious intent can easily use these open-sourced LLMs to 
generate harmful texts and dis/misinformation at scale. To mitigate this problem, a computational method to determine if a given text is a deepfake text or not is desired--i.e., Turing Test (TT). In particular, in this work, we investigate the more general version of the problem, known as \emph{Authorship Attribution (AA)}, in a multi-class setting--i.e., not only determining if a given text is a deepfake text or not but also being able to pinpoint which LLM is the author. 
We propose {\sc \textbf{TopFormer}} to improve existing AA solutions by capturing 
more linguistic patterns in deepfake texts by including 
a Topological Data Analysis (TDA) layer in the Transformer-based model. 
We show the benefits of having a TDA layer when dealing with imbalanced, and multi-style datasets, 
by extracting TDA features from the reshaped $pooled\_output$ of our backbone as input.
This Transformer-based model captures contextual representations (i.e., semantic and syntactic linguistic features), 
while TDA captures the shape and structure of data (i.e., linguistic structures). 
Finally,
{\sc \textbf{TopFormer}}, outperforms all baselines in all 3 datasets,
achieving up to 7\% increase in Macro F1 score. 
Our code and datasets are available at: \textcolor{blue}{\url{https://github.com/AdaUchendu/topformer}}
\end{abstract}

\end{frontmatter}

%%%%%%%%%%%%%%%%%%%%%%%%%%%%%%%%%%%%%%%%%%%%%%%%%%%%%%%%%%%%%%%%%%%%%%%%

\section{Introduction}

Recent Large Language Models (LLMs) now have a trillion parameters and 
are able to generate more human-like texts. 
These larger models pose a few difficulties, the more glaring being that reproducibility is very difficult and expensive. This allows LLMs to remain black-box, 
with their limitations maliciously exploited. 
Some of these limitations include: 
toxic and hate speech generation \cite{sheng2021societal}, 
plagiarism \cite{lee2022language}, 
memorization of sensitive information 
\cite{carlini2021extracting} and hallucinated text generation \cite{rawte2023survey}
which allow LLMs to be easily exploited to generate authentic-looking and convincing 
disinformation \cite{lucas2023fighting}.

\begin{figure}[tb]
    \centering
    \includegraphics[width=1 \linewidth]{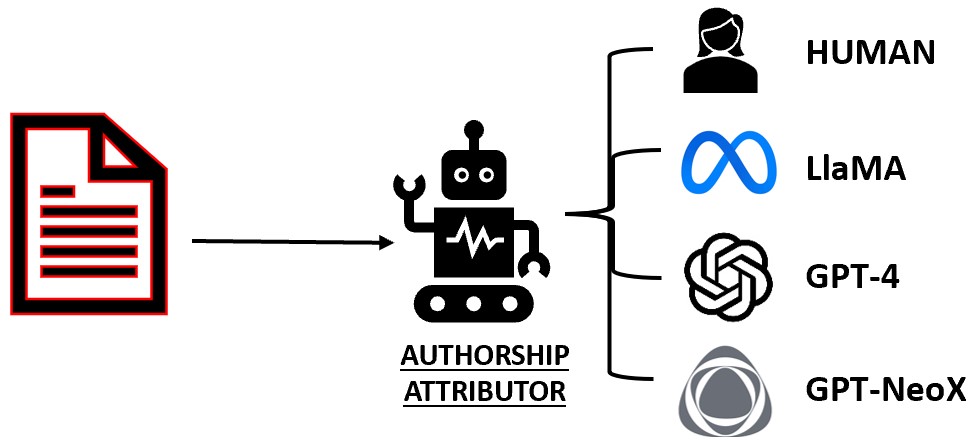}
    \caption{Illustration of the Authorship Attribution (AA) problem with multiple authors - human and many deepfake (LLM) authors. }
    \label{fig:aa_fig}
    \vspace{20pt}
\end{figure}

The first step to mitigate these limitations of LLMs starts with our ability to determine whether a text is generated by a particular LLM ({\bf deepfake text}\footnote{We call these LLM-generated texts, \textit{deepfake texts}; However, there are other terms which include - \textit{AI-generated, auto-generated, machine-generated, synthetic, artificial, computer-generated, neural, and LLM-generated texts}}) 
or not. This task is known as \emph{Turing Test (TT)}~\cite{uchendu2021turingbench}. Further, in this work, generalizing the TT problem further, we are interested in not only determining if a text is a deepfake text or not but also pinpointing which LLM is the author, known as the {\bf Authorship Attribution} (AA) problem~\cite{uchendu2022attribution}.
See the illustration of AA in Figure \ref{fig:aa_fig}.
The AA problem in a multi-class setting has not 
been as rigorously studied as the TT problem has. Naturally, AA is substantially more challenging than TT.
%due to its non-trivial nature. 
However, with the ever-increasing number of popular LLMs that people can use,
%usage of ChatGPT powered by GPT-3.5 or GPT-4, and other LLMs, 
we believe that it is no longer sufficient to just ask the binary question of ``is this written by human or machine?"
Solving the AA problem enables more fine-grained and targeted defense tools for users and platforms (e.g., a detector specifically trained and designed to detect ChatGPT deepfake texts).
AA solutions can also help researchers and policymakers understand the capacity and limitations of different LLMs,
and study which LLMs are more vulnerable to misuse and abuse in what context (e.g., political propaganda, terrorism recruitment, phishing). 

Researchers have proposed several solutions to distinguish deepfake texts from human-written texts, utilizing 
supervised and unsupervised machine learning \cite{uchendu2022attribution}. 
%\citet{uchendu2022attribution} surveys these solutions, creating taxonomies that will be discussed further in section \ref{related}. 
In the supervised learning setting, researchers have developed \textit{stylometric}, 
\textit{deep learning}, and \textit{hybrid} solutions for detecting
deepfake texts. Further, in 
the unsupervised learning setting,  \textit{statistical} solutions such as watermarking approaches \cite{kirchenbauer2023watermark}
have been developed. 
Intuitively, deep learning and hybrid-based techniques achieve the best performance in terms of accuracy. 
However, in terms of adversarial robustness, statistical-based (i.e., threshold-guided metric-based) techniques are the most robust models, with hybrid models taking second/first place in adversarial robustness \cite{uchendu2022attribution}. To that end, we propose a hybrid solution which is an ensemble of statistical and deep learning-based techniques to get both benefits - good performance and robustness. 
We hypothesize that if our model has adversarial robustness properties, it could also be noise-resistant and thus be robust to out-of-distribution and imbalanced datasets.

Thus, we propose {\sc \textbf{TopFormer}}, an ensemble of a Transformer-based (RoBERTa) model and Topological Data Analysis (TDA) techniques. 
We specifically choose
RoBERTa \cite{liu2019roberta} as the base model due to its state-of-the-art (SOTA) performance in feature extraction from text and its larger vocabulary compared to other transformer models such as BERT~\cite{devlin2018bert}. 
TDA is applied for deepfake text detection as it captures the true shape of data amidst noise \cite{port2018persistent,port2019topological,munch2017user,turkes2022effectiveness}. 
%Limited access to SOTA LLMs hampers dataset size, leading to imbalanced and noisy datasets. 
Limited access to SOTA LLMs leads to imbalanced and noisy datasets
which hamper the ability to train better deepfake text detectors.
The evolving nature of deepfake texts, influenced by instruction-tuned LLMs and diverse prompts \cite{deshpande2023toxicity}, poses challenges in authorship attribution. 
In addition, people use LLMs to do a mirage of tasks, including generation, paraphrasing, editing, and summarizing a piece of text. 
All these factors increase the variability of deepfake texts in our information ecosystem and the label imbalance between deepfake texts and human-written texts. 
Therefore, we hypothesize that {\sc \textbf{TopFormer}} can address this variability, achieving high performance on datasets reflecting current trends.

\section{Related Work} \label{related}

\subsection{Authorship Attribution of Deepfake Texts}
Since 2019, there have been several efforts to mitigate the malicious uses of LLM-generated texts (deepfake texts) by way of detection. As this task is non-trivial, 
different techniques have been attempted and can be split into - 
\textit{stylometric}, \textit{deep learning}, \textit{statistical-based}, 
and \textit{hybrid classifiers} (ensemble of two or more of the previous 
types) and more recently \textit{prompt-based}, as well as \textit{human-based approaches} 
\cite{uchendu2022attribution}. Furthermore, this task which is modeled as 
an Authorship Attribution (AA) task of human vs. either one or several 
deepfake text generators are studied as either binary 
(\textit{Turing Test}) or multi-class problem. The most popular is the binary class problem as there is more data for it than for multi-class. 

Thus, for the \textit{stylometric} solution, 
several solutions are using POS tags, lexical dictionaries, etc. to capture 
an author's unique writing style \cite{uchendu2020authorship, kumarage2023stylometric}. Next for the 
\textit{deep learning} solutions, researchers usually fine-tune 
BERT models and BERT variants 
\cite{uchendu2021turingbench, ippolito2020automatic, kumarage2023j,sotofew,abassy2024llm}. 
However, as fine-tuning can be computationally expensive and requires a lot of data which does not always exist, some researchers have proposed an unsupervised technique - \textit{statistical-based} solutions
\cite{gehrmann2019gltr,mitchell2023detectgpt}.
However, while deep learning-based classifiers perform very well, they are highly 
susceptible to adversarial perturbations \cite{uchendu2022attribution}. 
Therefore, researchers propose \textit{hybrid-based} solutions that fuse both 
\textit{deep learning} and \textit{statistical-based} or \textit{stylometric} solutions to improve 
performance and robustness \cite{kushnareva2021artificial, chen2023stadee}.
More recently, \textit{prompt-based} techniques are being used for 
attribution of deepfakes \cite{koike2023outfox, kumarage2023reliable,uchendu2023understanding}.
Lastly, for the \textit{human-based} approaches, to improve human detection of deepfake texts, researchers have utilized two main techniques - 
training \cite{gehrmann2019gltr, uchendu2023understanding}
and not training \cite{ippolito2020automatic}.

\begin{figure*}[htb!]
    \centering
    \includegraphics[width=1 \linewidth]{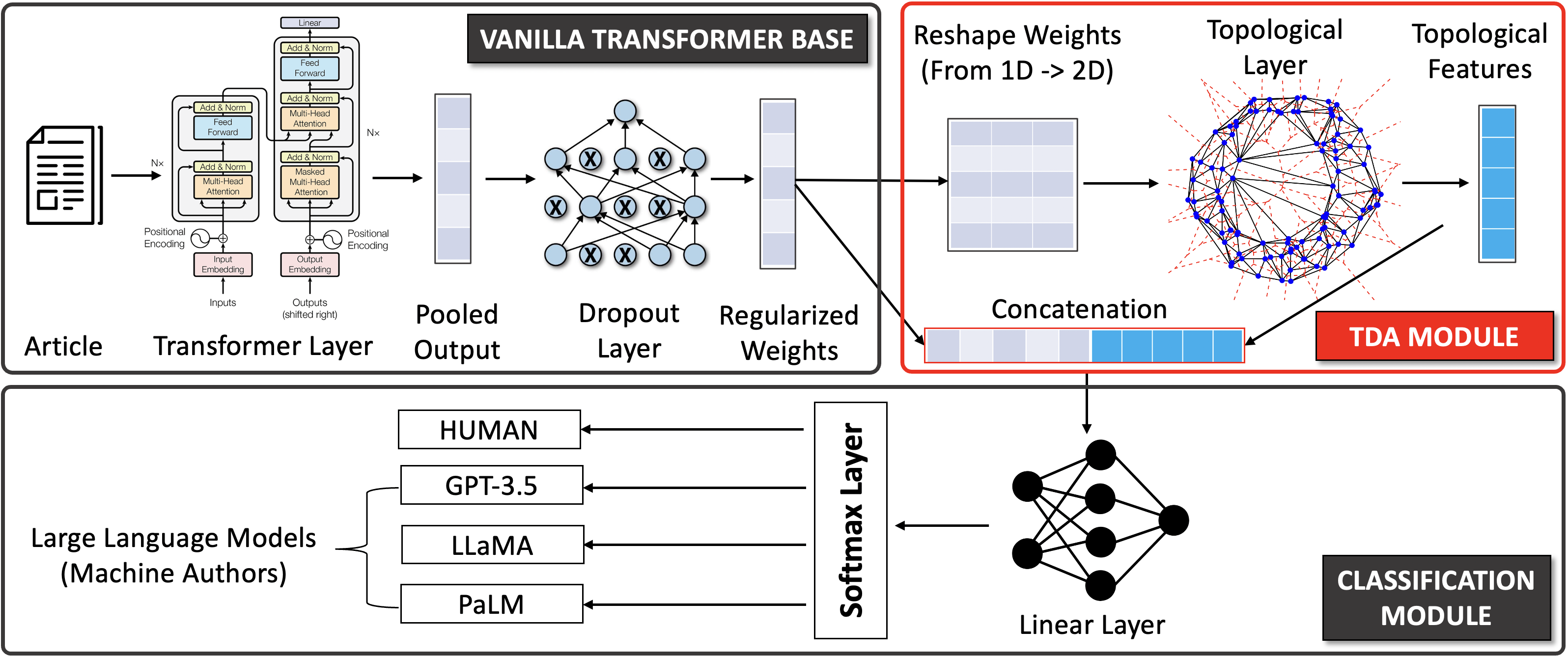}
    \caption{Flowchart of the Topological classification algorithm. The 
    \textcolor{red}{Red} frame indicates our methodology and technique to transform a Vanilla Transformer-based model to a Topological Transformer-based model.}
    \label{fig:flow}
        % \vspace{5mm}
    \vspace{5pt}
\end{figure*}

\subsection{TDA Applications in NLP}
\textbf{Topological Data Analysis (TDA)} is a technique often used to quantify shape and structure in data by leveraging concepts from algebraic topology. 
Due to this unique ability to obtain the true shape of data, in spite of noise, it has been 
implemented in machine learning problems. The NLP field has recently seen a recent uptake in TDA applications due to its benefits. TDA has been previously applied to 
a variety of Machine learning tasks 
\cite{hensel2021survey}, such as
detecting children and adolescent writing \cite{zhu2013persistent}, 
novelists attribution \cite{gholizadeh2018topological},
law documents analysis \cite{savle2019topological},
% time series analysis \cite{gholizadeh2020novel}
movie genre analysis \cite{doshi2018movie}, and
explanation of syntactic structures of different language families \cite{port2018persistent,port2019topological}. 
More recently, TDA techniques have been applied to 
the deepfake text detection problem \cite{kushnareva2021artificial}. However, they collect the 
statistical summaries of the TDA representations of BERT attention weights represented as a threshold-guided directed and undirected graph. 
Using these representations, they classify deepfake texts with Logistic regression
for the binary task - human vs. deepfake. 
% Therefore, for our technique, we train an end-to-end Transformer-based model with a TDA layer using the representations from the transformer-based model as the fine-tuning process continues. 
Next, \citet{perez2022topological} uses a similar technique as \citet{kushnareva2021artificial} 
to show that TDA can improve the robustness of BERT. 
Finally, TDA has also been applied to
representing documents as story trees \cite{haghighatkhah2022story}, 
% probing the contextualized embeddings of Transformer-based models \cite{rathore2023topobert},
detecting contradictions in texts \cite{wu2022topological}, 
examining the linguistic acceptability judgments of texts 
\cite{cherniavskii2022acceptability},
finding loops in logic \cite{tymochko2020argumentative},
speech processing \cite{tulchinskii2022topological}, 
% detecting fraudulent papers by examining their titles and abstracts \cite{tymochko2021connections}, 
and
extracting dialogue terms with Transfer learning
\cite{vukovic2022dialogue}.

\section{Topological Data Analysis (TDA) Features} \label{tda_background}
Topology is defined as ``the study of geometric properties 
and spatial relations unaffected by the continuous change 
of shape or size of figures,'' according to the Oxford Dictionary. 
Topological Data Analysis (TDA) is a ``collection of powerful tools that have the ability to quantify shape and structure in data''\footnote{\url{https://www.indicative.com/resource/topological-data-analysis/}}. There are two main TDA techniques - persistent
homology and mapper. We will only focus on persistent homology. 
Persistent homology is a TDA technique used to find topological patterns of the data \cite{tymochko2020argumentative}.

This technique takes in the data and represents it as a point
cloud, such that each point is enclosed by a circle. 
For this analysis, the aim is to extract the 
persistent features of the data using simplicial complexes. 
These formations extract features which are holes in different dimensions, 
represented as \textit{betti numbers} 
($\beta_d$, $d$-dimension). 
The holes in 0-Dimension ($\beta_0$),  
1-Dimension ($\beta_1$) and 2-Dimension ($\beta_2$), are 
called connected components, loops/tunnels, and voids, respectively. 

Finally, the TDA features recorded are the $birth$ (formation of holes), $death$ (deformation or the closing of holes), and persistence features in different dimensions. 
Persistence is defined as the length of time it took a feature to die ($death - birth$).
This means that if a point touches another point then one of the points/features has died. The $death$ is recorded with the radii value at which the points 
overlap. 
In addition, due to all the shifts and changes, from the 1-Dimension and upwards, some features may appear - a new hole, and this 
feature is recorded as a $birth$. The $birth$ feature is the radii at which it appeared. 

\begin{algorithm}[!htb]
\caption{{\sc \textbf{TopFormer}} Algorithm (Forward Pass)}
\begin{algorithmic}[1]
\STATE \textbf{Input:} Document $x$, pretrained transformer model $M(\cdot)$, fully-connected neural network $f(\cdot)$.
\STATE \textbf{Output:} Predicted authorship label $y'$
% \STATE \textbf{Step 1: Preprocessing}
% \STATE Tokenize the text data $D$ using the tokenizer associated with $M$
% \STATE Convert tokens to input embeddings for the Transformer model
% \STATE \textbf{Step 2: Contextual Representation Extraction}
% \STATE Pass the input embeddings through the Transformer model $M$
\STATE $x_\mathrm{pooled}$$\leftarrow$M(x) //Retrieve pooled output from $M$
\STATE $x_\mathrm{pooled}$$\leftarrow$reshape($x_\mathrm{pooled}$) //Reshape to a suitable form
\STATE $x_\mathrm{TDA}$$\leftarrow$TDA($x_\mathrm{pooled}$) //Extract topological features
\STATE $x^*$$\leftarrow$concatenate($x_\mathrm{pooled}$, $x_\mathrm{TDA}$)
\STATE $y'$$\leftarrow$softmax($f$($x^*$))
% \STATE \textbf{Step 3: Topological Data Analysis (TDA) Feature Extraction}
% \STATE Reshape the pooled output $P$ to a suitable form for TDA
% \STATE Apply TDA to $P$ to extract topological features $T$
% \STATE Concatenate $P$ and $T$ to form a combined feature vector $F$
% \STATE \textbf{Step 4: Classification}
% \STATE Pass $F$ through a fully connected layer followed by a softmax layer
% \STATE Obtain the predicted label $y'_i$ for each input text
% \STATE \textbf{Step 5: Training and Optimization}
% \STATE Compute the loss using cross-entropy between the predicted labels and the true labels
% \STATE Backpropagate the loss and update the model parameters
% \STATE Repeat until convergence
% \STATE \textbf{Step 6: Inference}
% \STATE For each new input text, repeat Steps 1-4 to obtain predicted labels
% \STATE \textbf{End Algorithm}
\STATE \textbf{Return:} $y'$
\end{algorithmic}
\label{algo}
\end{algorithm}

% \begin{figure}
%     \centering
%     \includegraphics[width=0.2 \textwidth]{figures/model.jpg}
%     \caption{Topological deepfake text detector architecture}
%     \label{fig:model}
% \end{figure}

\begin{figure*}[htb!]
    \centering
    \includegraphics[width=1.\linewidth]{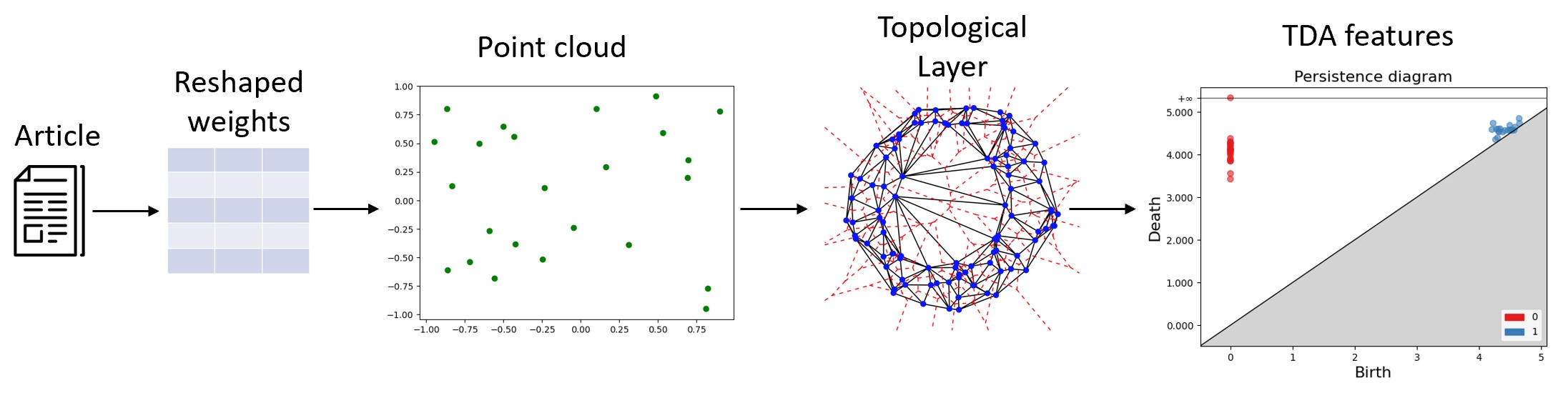}
    \caption{Illustration of how we extract the TDA features using the reshaped Transformer-based model's 
    regularized weights as input. First, we reshape the regularized $pooled\_output$ from $1 \times 768$ dimensions to $24 \times 32$
    and use this 2D matrix as input for the Topological layer. 
    The Topological layer treats this 2D matrix as a point cloud plot and 
    extracts TDA features ($birth$ \& $death$).
    Next, these TDA features are plotted in a figure known as 
    \textit{Persistent Diagram}, where the $birth$ features are 
    on the $x$-axis and $death$ features are on the $y$-axis. While we plot the features from the 0-Dimension (connected components) and 1-Dimension (loops), only 0-Dimension features are used for our task. }
    \label{fig:tda_feat}
        % \vspace{2mm}
    \vspace{10pt}
\end{figure*}

\section{{\sc \textbf{TopFormer}}: Topology-Aware Attributor} \label{method}

%\subsection{\textbf{TopRoBERTa\_{pool}}} \label{method-pool}
To build this TDA-infused Transformer-based model, we focus on the four layers 
needed to convert the vanilla Transformer-based (RoBERTa) model to a \textbf{Top}ological Trans\textbf{former}-based model ({\sc \textbf{TopFormer}})
- (1) pre-trained weights of the backbone model, 
(2) dropout layer with probability $p$=0.3, 
(3) Topological layer for calculating, and 
(4) Linear transformation layer. See Figure \ref{fig:flow} for a flow chart describing the architecture of {\sc \textbf{TopFormer}} with the 4 layers and Algorithm \ref{algo} for the pseudocode for ({\sc \textbf{TopFormer}}).

To train our end-to-end Topological model, 
we first fine-tune RoBERTa-base model. As we fine-tune the model,
we take the $pooled\_output$ which is a $1 \times 768$ vector containing the 
latent representations of the model from the final layer. 
We find that the weights are richer than BERT because it is a robustly trained BERT model and 
has over 20K more vocabulary size than BERT. 
These latent representations capture word-level and sentence-level 
relationships, thus extracting contextual representations \cite{liu2019roberta}. 
Due to the contextual representations captured, the weights essentially 
extract semantic and syntactic linguistic features \cite{tenney2019bert,reif2019visualizing}.

Next, we pass this $pooled\_output$ which is a 
$1 \times 768$ vector
into a regularization layer, called $dropout$. This $dropout$ layer drops 
a pre-defined percentage (30\% in our case) of our $pooled\_output$ to make our model 
more generalizable and less likely to overfit. 
This yields an output $dropout(pooled\_output)$ with the same dimensions as the input - $1 \times 768$ vector.

Before, we use our regularized output - $dropout(pooled\_output)$ as input for the 
Topological layer\footnote{\url{https://github.com/aidos-lab/pytorch-topological/tree/main}}, 
we first reshape it from 1D $\to$ 2D. TDA requires at least a 
2D matrix to construct simplicial complexes that persistent homology technique uses to 
extract the $birth$ and $death$ of TDA features (connected components, specifically) \cite{munch2017user}.
%\cite{munch2017user, hensel2021survey}.
This is because the simplicial complexes can only be extracted from 
the point cloud (which can be visualized as a scatterplot of the dataset) and 
to get this point cloud we need a dataset with 2-coordinates. 
We include this Topological layer in the backbone model because: 
(1) RoBERTa has richer latent representations than BERT \cite{liu2019roberta};
(2) TDA is robust to noise, out-of-distribution, and limited data \cite{turkes2022effectiveness};
(3) TDA is able to capture features that other feature extraction techniques cannot capture \cite{port2018persistent,port2019topological}; and
(4) TDA extracts the true structure of data points \cite{munch2017user}. 
To convert the regularized weights from 1D $\to$ 2D is non-trivial because we 
need to get the best shape to obtain useful TDA features which 
are stable and uniform (vectors of the same length) across 
all input of a particular dataset. 
Therefore, we tried different 2D sizes and found that the closer it is
to a dense square matrix, the more stable the TDA features are. 
Stable in this context means that for 
every input, the TDA layer outputs the same number of features in a vector. 
Therefore, we convert the $1 \times 768$ vector to a $24 \times 32$ matrix, since it 
is the closest to a square matrix as 768 is not a perfect square. 
We also find through experimentation that when 
row $>$ column, TDA features are unstable.
Unstable for our task means that the Topological layer output 
different vector sizes of TDA features given the input. Also, 
sometimes the feature vector can only contain $nan$ values 
based on the input which means that it was unable to extract 
TDA features. 
Thus, we find that $pooled\_output$ must be reshaped such that 
row $\leq$ column and $24 \times 32$ satisfies this claim.
Finally, using the 2D matrix as input to our Topological layer, we obtain the 
0-Dimension ($\beta_0$) features following the process illustrated in Section \ref{tda_background}. This yields a
$23 \times 3$. 
These 3 columns represent the $birth$ time, $death$ time, and persistence features, respectively.
Persistence is defined as the length of time it took a feature to die.
Next, this 2D matrix is 
flattened to a vector size of $1 \times 69$ so 
it can be easily concatenated with the 1D $dropout(pooled\_output)$. 
See Figure \ref{fig:tda_feat} for 
an illustration of how the TDA features are extracted. We interpret these TDA features as 
capturing linguistic structure, as it is capturing the structure and shape of textual data. 

Lastly, we concatenate the regularized backbone weights ($dropout(pooled\_output)$) of size
$1 \times768 $ with the TDA features of size $1 \times 69$. 
This 
yields a vector of size $1 \times 837$. Thus, this $1 \times 837$ vector serves as 
input for the final layer of feature transformation, the Linear layer. 
The Linear layer's latent space increases from 
$768$ to $837$ in order to take the concatenated vector as input. 
TDA increases the latent space by $69$ dimensions. However, we observe that unlike other TDA-based Transformer classifiers \cite{kushnareva2021artificial,perez2022topological},
which use attention weights in which the size is dependent on the length of text,
our TDA technique increases the latent space minimally. 
Finally, the output of this Linear layer is a 
vector that is the size of $batch\_size \times number\_of\_labels$. 
Thus, if we have $batch\_size=16$ and $number\_of\_labels=20$, we obtain a vector of size: 
$16 \times 20$. Finally, we pass this vector as input 
into the softmax layer for multi-class classification. 

% We train this Topological model for 5 epochs so both the 
% RoBERTa pre-trained weights and Topological features can be improved. 
% We use Adam optimizer, cross-entropy loss, and a learning rate of $2e-5$.
% We experimented with different loss functions - cross-entropy, topological loss, 
% contrastive loss, Gaussian loss, and different combinations of these loss functions. 
% We find that cross-entropy loss achieves the best performance.
Finally, TDA features are compatible with non-TDA features \cite{mcguire2023neural}, 
making it a suitable technique for extracting subtle linguistic patterns that 
distinguish deepfake texts from human-written ones. 
Thus, {\sc \textbf{TopFormer}} captures semantic, syntactic, and structural linguistic features. 

%\vspace{-3mm}

\section{Experiments}

\subsection{Datasets}
% Since this paper focuses on multi-class authorship attribution of deepfake texts vs. human-written texts, we evaluated our model on three deepfake text datasets. 
% Furthermore, these datasets are a reflection of the real world, where we currently have more human-written text examples than deepfake texts. 

\subsubsection{OpenLLMText}
OpenLLMText \cite{chen2023token} dataset was collected from five authors - 
\textit{human, LLaMA, ChatGPT, PALM,} and \textit{GPT-2}. For all the LLaMA, ChatGPT, and PALM labels, the text was generated by paraphrasing the human-written texts.
%LLaMA was prompted to generate realistic-looking texts similar to human-written texts without paraphrasing. 
%Similar to LLaMA, 
GPT-2 was prompted to completely generate new texts given some of the human-written texts in the prompt. 
%This dataset utilizes different prompting strategies for generation, making a suitable dataset for our task. 
Thus, due to the different NLG methods employed, this data also has 3 writing style labels - human, paraphrasers, and generators. 
However, to make the dataset further reflect the real world where there is gross data imbalance (human label $\geq$ machine label), 
we take 1\% of all the deepfake labels. And we could do this because the initial dataset had 340K samples. 
%This yielded 7367 human labels and 73 deepfake labels for each LLM. 
See Table \ref{tab:data} for the train, validation, and test splits.

\begin{table}[!htb]
    \centering
    \footnotesize
    % \resizebox{7cm}{!}{
    \begin{tabular}{ccccc}
        \toprule
     \textbf{Dataset}    & \textbf{Train} & \textbf{Valid} & \textbf{Test} & 
     \textbf{\# Labels} \\%& \textbf{\# Style Labels}\\
     \cmidrule(lr){1-5}
        OpenLLMText & 53K & 10K & 7.7K & 5\\
        % \hline
        SynSciPass & 87K & 10K & 10K & 12  \\
        Mixset & 2.4K & 340 & 678 & 8 \\
        \bottomrule 
    \end{tabular}
    % }
    \caption{Dataset summary statistics.}
    \label{tab:data}
    % \vspace{10pt}
\end{table}

\subsubsection{SynSciPass}
SynSciPass \cite{rosati2022synscipass} dataset is comprised of scientific articles, 
authored, by both human and deepfake authors. 
In addition to being grossly imbalanced (i.e., 79K examples for human \& 600-850 examples for deepfake labels). 
This is because the
deepfake texts are generated with open-ended text-generators like GPT-3, SynSciPass's deepfake labels are generated with 3 types of text-generators. These are open-ended generators, translators like Google translate (e.g. English $\to$ Spanish $\to$ English), and paraphrasers like SCIgen and Pegasus.
Using these different text-generation techniques introduces high variance in writing styles in 
this dataset.
We use the 12 labels - 1 human \& 11 deepfake text-generators. 
However, due to the different NLG methods employed, this data also has 4 writing style labels - human, generators, translators, and paraphrasers. 
See Table \ref{tab:data} for the train, validation, and 
test splits.
%Finally, since this dataset looks at the task from a different perspective, as well as being grossly imbalanced, we evaluated our models on these constraints. 

\subsubsection{Mixset}
The Mixset \cite{gao2024llm} dataset has 3 labels - human, machine, and mixset. 
Machine labels are for the texts that were fully generated with the LLMs, not 
paraphrased, while mixset is for texts that were paraphrased either by humans or 
different LLMs using diverse paraphrasing prompts to paraphrase both the human-written texts and LLM-generated texts. 
We use 8 labels for the distinct authorships - \textit{human, GPT-4, LLaMA, Dolly, ChatGLM, StableLM, ChatGPT-turbo}, and \textit{ChatGPT}. Only LLaMA and GPT-4
were used for paraphrasing, while ChatGLM, Dolly, StableLM, ChatGPT, and ChatGPT-turbo are used for generation (i.e., generating entirely new texts). 
Thus, due to the different NLG methods employed, this data also has 3 writing style labels - human, paraphrasers, and generators. 
Finally, to keep in the context of real-world settings, due to the small size of the data, we were only able to get sufficient samples per label with as low as 10\% of the non-human (deepfake) labels. 
See Table \ref{tab:data} for the train, validation, and test splits.

\subsection{Authorship Attribution Models} \label{author_attrib}
We evaluate a diverse set of authorship attribution models:

\begin{itemize}[itemsep=3pt,topsep=0pt]

    \item \textbf{GPT-who} \cite{venkatraman2023gpt}: 
    is a psycholinguistically-aware domain-agnostic statistical-based deepfake text detector. It obtains token probabilities using GPT-2 XL pre-trained weights and calculates the uniform information density features and uses Logistic regression for classification.

     % \item \textbf{OpenAI detector}\footnote{https://huggingface.co/openai-community/roberta-base-openai-detector}: We use roberta-base-openai-detector pre-trained model. This model is RoBERTa-base fine-tuned by Huggingface to detect GPT-2 texts. However, it can be repurposed for AA of deepfake authors. It serves as a good baseline in the \textit{deepfake text detection} niche community. 

     \item \textbf{Contra-BERT} \cite{ai2022whodunit}:
     is a BERT-base model that is trained with contrastive and cross-entropy loss functions. \citet{ai2022whodunit} shows that the model can achieve SOTA performance when applied to both traditional AA (i.e., only human authors) and deepfake text attribution.

    \item \textbf{BERT:} We use BERT-base cased pre-trained model. This is also typically used as a strong baseline for deepfake text attribution (often outperforming proposed detectors) \cite{uchendu2022attribution}. 

     %\item \textbf{TopBERT\_{attn}:} We add a Topological layer to the BERT model described above and follow the process described in Section \ref{method-attn}.

    % \item \textbf{TopBERT:} We add a Topological layer to the BERT model described above and follow the process described in Section \ref{method}
    % and Figure \ref{fig:flow}.

    % \item \textbf{Gaussian-BERT:} A BERT-base model with a Gaussian layer to add Gaussian noise to the weights. The hypothesis is that if TopBERT achieves superior performance randomly then adding a Gaussian layer should have a similar effect. 

    \item \textbf{RoBERTa:} We use RoBERTa-base pre-trained model. This is another typical strong baseline for deepfake text attribution (often outperforming proposed detectors) \cite{uchendu2022attribution}. 

    %\item \textbf{TopRoBERTa\_{attn}:} We add a Topological layer to the RoBERTa model and follow the process described in Section \ref{method-attn}.

        \item \textbf{Gaussian-RoBERTa:} RoBERTa-base model with a Gaussian layer to add Gaussian noise to the weights. The hypothesis is that if {\sc \textbf{TopFormer}} achieves superior performance randomly, then adding a Gaussian layer should have a similar effect. 
        
    %The hypothesis also remains the same here but applied to TopRoBERTa.

    \item \textbf{{\sc \textbf{TopFormer}}:}  RoBERTa-base with a Topological layer, following the process described in Section \ref{method} and Figure \ref{fig:flow}.

\end{itemize}

We train all the models, except GPT-who and Contra-BERT with the same hyperparameters \& parameters - MAX\_length = 512, dropout probability $p=0.3$, learning rate of 2e-5, cross-entropy loss, batch size of 16 and
5 epochs. 
Also, we tested our {\sc \textbf{TopFormer}} model with other loss functions (contrastive loss, topological loss, and Gaussian loss) and found cross-entropy to be the best.
Lastly, we evaluate these models using 
established evaluation metrics for machine learning - 
Precision, Recall, Accuracy, and Macro F1 score. 
%However, since the datasets are imbalanced, we focus on the Macro F1 score which we use to calculate the percentage gains for the classification task.

\newcolumntype{H}{>{\setbox0=\hbox\bgroup}c<{\egroup}@{}}
\renewcommand{\tabcolsep}{5pt}
\begin{table}[tb]%[ht]
    \centering
     % \resizebox{11cm}{!}{
    \begin{tabular}{ccccHcH}
      \toprule
      \textbf{MODEL}   &  \textbf{Precision} & \textbf{Recall} & \textbf{Accuracy}  & \textbf{Weighted F1} & \textbf{Macro F1} & \textbf{\% Gain}\\
      \cmidrule(lr){1-7}
      GPT-who & 0.1924 & 0.2000 & 0.9619 & 0.9432 & 0.1961 \\
      % OpenAI detector  \\
      Contra-BERT & \textbf{0.8112} & 0.6587 & 0.9761 & 0.9736 & 0.6696\\
      % \cmidrule(lr){1-7}
       BERT   &   0.7914 & 0.7190 & \textbf{0.9792} & \textbf{0.9792} & 0.7255 & - \\ 
       % \hline 
      % Gaussian-BERT & 0.7917 & 0.6500 & 0.9734 & 0.9722 & 0.6607 & 
      % \textcolor{red}{ 6\% $\downarrow$} \\
    
       % % \hline
       %  TopBERT & \textbf{0.8246} & 0.6588 & \underline{0.9768} & 0.9742 & 0.6786 & 
       %  \textcolor{red}{ 4\% $\downarrow$} \\
        
       % \cmidrule(lr){1-7}
       RoBERTa &  \underline{0.8099} & \underline{0.7428} & 0.9764 & 0.9763 & \underline{0.7288} & - \\
       
       Gaussian-RoBERTa & 0.6231 & 0.5009 & 0.9740 & 0.9674 & 0.5321 &
       \textcolor{red}{ 20\% $\downarrow$}\\
        
        \cmidrule(lr){1-7}

        {\sc \textbf{TopFormer}} & {0.8069} & \textbf{0.7727 }& \underline{0.9766} & \underline{0.9774} & \textbf{0.7522} & 
        \textcolor{blue}{ 2\% $\uparrow$}\\
       \bottomrule 
      
    \end{tabular}
    % }
        \caption{OpenLLMText Authorship Attribution results. The best performance is \textbf{boldened} and the second best is \underline{underlined}. 
        % The percentage gains reported in the \textit{\% Gain} in Macro F1.
        }
    \label{tab:llm_results}
        % \vspace{5mm}
        \vspace{10pt}
\end{table}

\renewcommand{\tabcolsep}{5pt}
\begin{table}[tb]%[!htbp]
    \centering
    % \resizebox{11cm}{!}{
    \begin{tabular}{ccccHcH}
      \toprule
      \textbf{MODEL}   &  \textbf{Precision} & \textbf{Recall} & \textbf{Accuracy} & \textbf{Weighted F1} & \textbf{Macro F1}  & \textbf{\% Gain} \\
      \cmidrule(lr){1-7}
      GPT-who & 0.2224 & 0.1434 & 0.9126 & 0.8793 & 0.1622  \\
      % OpenAI detector & \\
      Contra-BERT & 0.7967 & 0.7307 & 0.9703 & 0.9686 & 0.7492 \\
      % \cmidrule(lr){1-7}
       BERT   & 0.8585 & 0.8148 & 0.9791  & 0.9785 &  0.8327  & - \\
       % \hline 
        % Gaussian-BERT &  0.8404 & 0.7709 & 0.9745  & 0.9735 & 0.7933 & \textcolor{red}{ 4\% $\downarrow$}\\
     
         % TopBERT & 0.8682 & 0.8298 & 0.9807  & 0.9802 & 0.8471 & 
         % \textcolor{blue}{ 2\% $\uparrow$} \\
       % \hline
       % \cmidrule(lr){1-7}
       RoBERTa &  \underline{0.9012} &  0.8554 & 0.9853 & 0.9846 & 0.8719   & - \\

       Gaussian-RoBERTa &  0.8929 & \underline{0.8809} & \underline{0.9872}  & \underline{0.9870} & \underline{0.8847} & \textcolor{blue}{ 1\% $\uparrow$} \\

        \cmidrule(lr){1-7}
       
        {\sc \textbf{TopFormer}} &  \textbf{0.9177} & \textbf{0.8978 }& \textbf{0.9892}  & \textbf{0.9890} & \textbf{0.9058 } &  \textcolor{blue}{ 4\% $\uparrow$}\\
       \bottomrule 
       
    \end{tabular}
    % }
        \caption{SynSciPass Authorship Attribution results. The best performance is \textbf{boldened} and the second best is \underline{underlined}. 
        % The percentage gains reported in the \textit{\% Gain} in the Macro F1.
        }
    \label{tab:syn_results}
        % \vspace{5mm}
                \vspace{10pt}
\end{table}

\renewcommand{\tabcolsep}{5pt}
\begin{table}[tb!]
    \centering
     % \resizebox{11cm}{!}{
    \begin{tabular}{ccccHcH}
      \toprule
      \textbf{MODEL}   &  \textbf{Precision} & \textbf{Recall} & \textbf{Accuracy} & \textbf{Weighted F1} & \textbf{Macro F1}  & \textbf{\% Gain}\\
      \cmidrule(lr){1-7}
      GPT-who & 0.2825 & 0.2446 & 0.6647 & 0.5896 & 0.6647 \\
      % OpenAI detector & 0.6529 & 0.6673 & 0.8588 & 0.8537 & 0.6494 \\
      Contra-BERT &  0.7338 & 0.7411 & 0.8882 & 0.8809 & 0.7287 \\
      % \cmidrule(lr){1-7}
       BERT   & \underline{0.7982} & \underline{0.8214} & \underline{0.9118} & \underline{0.9034} & \underline{0.8034} & - \\ 
       % \hline 
      % Gaussian-BERT  & 0.5614 & 0.5875 & 0.8265 & 0.8115 & 0.5556 & \textcolor{red}{ 25\% $\downarrow$} \\
  
        % TopBERT  & \textbf{0.8541} & \textbf{0.8631} & \textbf{0.9324} & \textbf{0.9261} & \textbf{0.8480} &
        % \textcolor{blue}{ 5\% $\uparrow$} \\
       % \hline
       % \cmidrule(lr){1-7}
       RoBERTa & {0.7697} & 0.7976 & 0.9000 & 0.8927 & 0.7705 & - \\
       % \hline 
        Gaussian-RoBERTa & 0.4014 & 0.3862 & 0.7404 & 0.7412 & 0.7404 & \textcolor{red}{ 3\% $\downarrow$} \\
        
       \cmidrule(lr){1-7}

        {\sc \textbf{TopFormer}} & \textbf{0.8181} & \textbf{0.8268} & \textbf{0.9176} & \textbf{0.9110} & \textbf{0.8294} & 
        \textcolor{blue}{ 6\% $\uparrow$}\\
       \bottomrule 
      
    \end{tabular}
    % }
        \caption{Mixset Authorship Attribution results. The best performance is \textbf{boldened} and the second best is \underline{underlined}. 
        % The percentage gains reported in the \textit{\% Gain} in the Macro F1.  
        }
    \label{tab:m_results}
        % \vspace{5mm}
        \vspace{10pt}
\end{table}

\section{Results}

Our proposed model - {\sc {TopFormer}}
is 
evaluated on 
its ability to more accurately attribute human- vs. deepfake-authored articles 
to their true authors. 
We specifically used imbalanced multi-style datasets  - OpenLLMText, SynSciPass, and 
Mixset to simulate the current landscape of deepfake texts vs. human-written texts in the real world. 
From Tables \ref{tab:llm_results}, \ref{tab:syn_results}
and \ref{tab:m_results},
we observe that {\sc {TopFormer}} excels in the AA task, consistently outperforming the baseline SOTA deepfake AA models in all 3 datasets.  

{\sc {TopFormer}} outperforms in all 3 datasets, 
achieving the highest Macro F1 scores. 
Additionally, the customized baseline deepfake attribution models - GPT-who and Contra-BERT underperform by a large margin. 
However, Gaussian-RoBERTa underperforms RoBERTa in all datasets, except for SynSciPass which minimally outperforms by 1\%. 
Other deepfake attribution baseline models - BERT \& RoBERTa 
underperform as well, with BERT outperforming RoBERTa in only the Mixset 
dataset, possibly because of the small nature of the dataset. 

In addition, adding a TDA layer to RoBERTa increases the training time and inference time by about 0.5-2 hours, depending on data size. However, just inference time remains fast.

% For the OpenLLMText dataset, we observe that  {\sc \textbf{TopFormer}} achieved the highest Macro F1 score (75\%), with Gaussian-RoBERTa underperforming 
% both BERT and RoBERTa. 
% However, for both SynSciPass and Mixset datasets, 
% we observe that {\sc \textbf{TopFormer}}
% outperformed all baseline models in all the metrics. 
% In addition, Gaussian-RoBERTa also improved in performance minimally by 1\%
% for the SynSciPass dataset.
% Lastly, for all 3 datasets the more customized deepfake AA models - GPT-who and Contra-BERT underperform by a large margin, suggesting that both this task  
% and the improvements witnessed in {\sc \textbf{TopFormer}} are non-trivial. 

% Where TopFormer does not work 

\section{Further Analysis}

\renewcommand{\tabcolsep}{2.7pt}
\begin{table}[bt]
    \centering
    \footnotesize
    % \resizebox{12cm}{!}{
    \begin{tabular}{ccccccc}
      \toprule
       \multicolumn{1}{c}{} & \multicolumn{2}{c}{\textbf{OpenLLMText}} 
       & \multicolumn{2}{c}{\textbf{SynSciPass}} & 
       \multicolumn{2}{c}{\textbf{Mixset}} \\
      \cmidrule(lr){2-3} \cmidrule(lr){4-5} \cmidrule(lr){6-7}
      \textbf{MODEL} &  
      \textbf{Macro F1} & \textbf{\%$\Delta$} &
      \textbf{Macro F1} & \textbf{\%$\Delta$} & 
      \textbf{Macro F1} & \textbf{\%$\Delta$} \\
      \cmidrule(lr){2-7}
      % BERT & 0.6837 & - & 0.9414 & - & 0.8976 & \\ 
      % {\sc \textbf{TopFormer}} w/ BERT & 0.6958 & \textcolor{blue}{ 2\% $\uparrow$}  
      % & 0.9412 & \textcolor{black}{0\%} & 0.9379 
      % & \textcolor{blue}{ 4\% $\uparrow$} \\
      % \cmidrule(lr){1-7}
      RoBERTa & 0.7320 & - & 0.9064 & - &  0.9255 & - \\
      {\sc \textbf{TopFormer}} & 0.7331 & \textcolor{black}{ 0\%} 
      & 0.9746 & \textcolor{blue}{7\% $\uparrow$} 
      & 0.9240 & \textcolor{black}{0\%} \\
      \bottomrule 
    \end{tabular}
    % }
    \caption{Deepfake Text Style Detection results. \%$\Delta$ is the \% Gains in Macro F1.}
    \label{tab:style}
        % \vspace{5mm}
        \vspace{10pt}
\end{table}

\renewcommand{\tabcolsep}{1pt}
\begin{table}[bt]
    \centering
    \footnotesize
    % \resizebox{12cm}{!}{
    \begin{tabular}{ccccccc}
      \toprule
       \multicolumn{1}{c}{} & \multicolumn{2}{c}{\textbf{OpenLLMText}} 
       & \multicolumn{2}{c}{\textbf{SynSciPass}} & 
       \multicolumn{2}{c}{\textbf{Mixset}} \\
      % \toprule
      \cmidrule(lr){2-3} \cmidrule(lr){4-5} \cmidrule(lr){6-7}
      \textbf{MODEL} &  
      \textbf{Macro F1} & \textbf{\%$\Delta$} &
      \textbf{Macro F1} & \textbf{\%$\Delta$} & 
      \textbf{Macro F1} & \textbf{\%$\Delta$} \\
      \cmidrule(lr){2-7}
      % BERT & 0.6837 & - & 0.9414 & - & 0.8976 & \\ 
      % {\sc \textbf{TopFormer}} w/ BERT & 0.6958 & \textcolor{blue}{ 2\% $\uparrow$}  
      % & 0.9412 & \textcolor{black}{0\%} & 0.9379 
      % & \textcolor{blue}{ 4\% $\uparrow$} \\
      % \cmidrule(lr){1-7}
      BERT & 0.7255 & - & 0.8327 & - &  0.8034 & - \\
      Gaussian-BERT & 0.6607 & \textcolor{red}{ 6\%$\downarrow$}  & 
      0.7933 & \textcolor{red}{ 4\%$\downarrow$}  & 0.5556 &
      \textcolor{red}{ 24\%$\downarrow$}  \\
      {\sc \textbf{TopFormer}} w/ BERT & 0.6786 & \textcolor{red}{ 4\%$\downarrow$} 
      & 0.8471 & \textcolor{blue}{2\% $\uparrow$} 
      & 0.8480 & \textcolor{blue}{5\% $\uparrow$}  \\
      \bottomrule 
    \end{tabular}
    % }
    \caption{Results from using BERT as the Transformer-based backbone. \%$\Delta$ is the \% Gains in Macro F1.}
    \label{tab:bert}
        \vspace{10pt}
\end{table}

\subsection{Deepfake Text Style Detection}
To further show the utility of {\sc {TopFormer}} in the ever-evolving LLM landscape, 
we evaluated our model on the different writing styles of the current LLMs. 
Initially, users would only prompt the models with some of the text and the models would generate the rest. However, due to more recent improvements in NLG, LLM users, could create LLM-generated texts through paraphrasing, open-ended generation, summarization, translation, etc. Therefore, it is not only important to know which LLM authored a piece of text, but it could also be beneficial to figure out if it is through translation, paraphrasing, or open-ended generation. This knowledge, given the context and domain, can help ascertain the intention of the user - whether to improve their writing, plagiarize, evade detection, 
generate disinformation, etc. 

We evaluate OpenLLMText and Mixset on 3 labels - human, paraphrasers, and generators, while SynSciPass has 4 labels - human, paraphrasers, translators, and generators. 
See Table \ref{tab:style} for results. 
We observe that {\sc {TopFormer}} only outperforms when evaluated on the SynSciPass dataset, improving by 7\%. This could be because the SynSciPass dataset has one more writing style than the other datasets and TDA seems to thrive on high variability. 
Additionally, {\sc {TopFormer}} performs comparably to RoBERTa on the other 2 datasets - OpenLLMText and Mixset.
This suggests that TDA-based techniques can also benefit deepfake text style detection, especially with high style variability.

\subsection{Alternative Transformer-based Backbone}
We choose RoBERTa as our backbone because it remains the SOTA or at least a decent baseline for text classification tasks, including deepfake text attribution \cite{uchendu2022attribution}. 
However, we wanted to investigate how well a different backbone - 
BERT will perform when a Topological layer is included in the model. 
We compare vanilla BERT to {\sc {TopFormer}} w/ BERT. 
In addition, we compare these models to Gaussian-BERT to show 
{\sc {TopFormer}}'s performance is not related to being trained 
with noise but extracting useful features. 
See results in Table \ref{tab:bert}. 

{\sc {TopFormer}} w/ BERT outperforms when evaluated on the SynSciPass and 
Mixset datasets (by a 2\% and 5\% increase). 
While underperforming for OpenLLMText dataset by a 4\% decrease. {\sc {TopFormer}} w/ BERT performed the best on the smallest dataset - Mixset, suggesting that it may be more suitable for smaller datasets. 
Additionally, Gaussian-BERT consistently underperformed by large margins, suggesting that {\sc {TopFormer}} w/ BERT's performance is not due to training with noise but is learning and extracting useful (additional) features. 
Finally, these results suggest that including a Topological layer in a Transformer-based model could increase the performance depending on the heterogeneity of the dataset.

\subsection{Homogeneous Deepfake Text Benchmarks}
We claim that TDA improves performance when evaluated on 
datasets that better reflect the current landscape - multi-style deepfake texts.
To show where TDA does not work so well,
we evaluate {\sc {TopFormer}} on the more homogeneous datasets, 
where there are only 2 styles - human and one style of deepfake text generation. 
For a fairer comparison, we also skewed each dataset to have only 10\% of deepfake labels. We use TuringBench (TB) \cite{uchendu2021turingbench} which has 20 labels (i.e., 1 human \& 19 deepfake labels) and Monolingual English M4 \cite{wang2023m4} which has 6 labels (i.e, 1 human \& 5 deepfake labels). Table \ref{tab:error} shows that {\sc {TopFormer}} marginally underperforms and outperforms on the TB and M4 datasets by 1\% for both. 
This is because of the homogeneity of the dataset, however, we can still observe that {\sc {TopFormer}} performed comparably to the base model - RoBERTa, 
suggesting that even when TDA fails there is no significant loss of performance.

\newcolumntype{H}{>{\setbox0=\hbox\bgroup}c<{\egroup}@{}}
\renewcommand{\tabcolsep}{6.3pt}
\begin{table}[tb!]
    \centering
    \begin{tabular}{lccccHcH}
      \toprule
      {} & \multirow{2}{*}{\textbf{MODEL}} & \multicolumn{6}{c}{\textbf{METRICS}}\\
      \cmidrule(lr){3-7}
      {}   & {}   &  \textbf{Pre} & \textbf{Rec} & \textbf{Acc}  & \textbf{W.F1} & \textbf{Mac.F1} & \textbf{\% Gain}\\
       \multirow{3}{*}{\begin{sideways}TB\end{sideways}}  
       
       % & BERT   &  0.7247 & 0.7202 & 0.8115& 0.8083 & 0.7152  &  -  \\ 
       % \hline 
      % {}   & Gaussian-BERT  & 0.6987 & 0.7013 & 0.7964 & 0.7947 & 0.6975  &
      % \textcolor{red}{ 1\% $\downarrow$} \\
    
       % % \hline
       %  {}   & TopBERT &  0.7394  &  0.7358 & 0.8219 &  0.8177 & 0.7300   &
       %  \textcolor{blue}{ 1\% $\uparrow$}\\
        
       {}   & RoBERTa & \textbf{0.7555} & \textbf{0.7505} & \textbf{0.8333}  & \textbf{0.8275} & \textbf{0.7433} & - \\
        {}   & Gaussian-RoBERTa & 0.6500  &  0.6487  &  0.7609    &  0.7588 & 0.6452 &
        \textcolor{red}{ 9\% $\downarrow$}  \\
      
        {}   & {\sc \textbf{TopFormer}}  & \underline{0.7499} &  \underline{0.7452}  & \underline{0.8283}    & \underline{0.8200} & \underline{0.7313} &
        \textcolor{red}{ 1\% $\downarrow$} \\
      % {} & XXX & \\
       % \cmidrule(lr){1-7}
        % {}   &{\sc \textbf{TopFormer}} & 

       \cmidrule(lr){1-7}
        \multirow{3}{*}{\begin{sideways}M4\end{sideways}}   
        
        % &
      %   BERT   &  0.8628  & 0.9082  &  0.9347 & 0.9362 & 0.8814  &  -  \\ 
      %  % \hline 
      %  {}   & Gaussian-BERT  &  0.8341  &  0.8955  &  0.9164  & 0.9198 & 0.8580  &
      % \textcolor{red}{ 2\% $\downarrow$} \\
  
        %  {}   & TopBERT  &  0.8548  &  0.9052  &  0.9307  & 0.9319 &  0.8764   &
        % \textcolor{red}{ 0.5\% $\downarrow$}\\
       % \hline
        {}   & RoBERTa & \underline{0.9657}  &  0.9681  &  \underline{0.9852} &  \underline{0.9853} &  0.9656   & - \\
       % \hline 
         {}   & Gaussian-RoBERTa & 0.9644  &  \underline{0.9719}  &  0.9844  &  0.9845 &  \underline{0.9673}    &
        \textcolor{blue}{ 0.2\% $\uparrow$}  \\

         {}   & {\sc \textbf{TopFormer}} & \textbf{0.9745}  &  \textbf{0.9762}  &  \textbf{0.9893}    & \textbf{0.9893} & \textbf{0.9748} &
        \textcolor{blue}{ 1\% $\uparrow$} \\
       \bottomrule 
      
    \end{tabular}
    
    \caption{%\textbf{Error Analysis}: 
    Authorship Attribution results in Precision (Pre), Recall (Rec), Accuracy (Acc), and Macro F1 (Mac.F1) on TuringBench (TB) and M4. 
    The best and second best is \textbf{boldened} and \underline{underlined}. 
    % The percentage gains reported in the \textit{\% Gain} are calculated from the Macro F1.
    }
    \label{tab:error}
        % \vspace{5mm}
    \vspace{10pt}
\end{table}

\section{Ablation study}
We compare different inputs to the Topological layer as follows.
\begin{enumerate}%[itemstep=3pt]

     \item \textbf{Pooled\_output}: Our technique uses the reshaped 2D matrix of the $pooled\_output$ as input. We call this model, \textit{{\sc \textbf{TopFormer}} (Ours)}. 
 
    \item \textbf{Attention weights}: We use the attention 
    weights which is of size 
    $Max\_length \times 768$
    as input for the Topological layer. 
    This technique is inspired by \citet{kushnareva2021artificial,perez2022topological}, 
    who use threshold-guided directed and undirected graphs of the attention weights for binary classification. Thus, instead of increasing the computational cost by building graphs with the attention weights, 
    we use the attention weights as input for extracting the TDA features. 
    This technique provides a fairer comparison to 
    \textit{{\sc {TopFormer}} (Ours)}. We will call the model using the attention weights - \textit{{\sc \textbf{TopFormer}}\_{attn}}.

    \item \textbf{Last\_hidden\_state weights}: We use the last hidden state that is a 3D matrix with size $12 \times Max\_length \times Max\_length$ as input for the TDA layer. This increases the size of the Linear latent space from 768 to 19,164.
    We call this model - \textit{{\sc \textbf{TopFormer}}\_{hidden}}

    \item \textbf{Correlation of Pooled\_output}: This is another intuitive technique where we multiply the transposed $pooled\_output$ vector (with size $768 \times 1$) to the $pooled\_output$ (with size $1 \times 768$) to obtain a 2D matrix, creating a correlation matrix of the vector. This yields a $768 \times 768$ matrix which is input for the TDA layer, increasing the size of the Linear latent space from 768 to 3068.
    We call this model - \textit{{\sc \textbf{TopFormer}}\_{pool\_corr}}.
    
\end{enumerate}

\begin{figure*}%[!htb]
\centering
\subfloat[RoBERTa-SynSciPass.]{\label{fig:rob_sci} 
\includegraphics[width=0.235\textwidth]{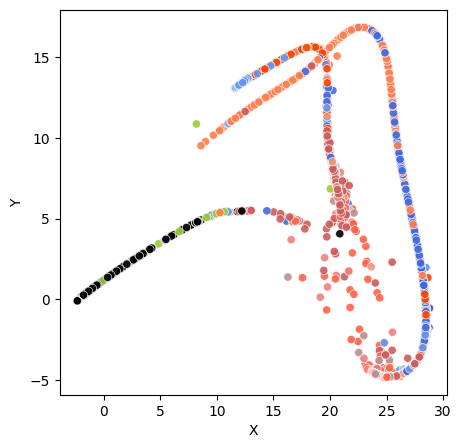}}
\hfill
\subfloat[{\sc \textbf{TopFormer}}-SynSciPass.]{\label{fig:toprob_sci} 
\includegraphics[width=0.2351\textwidth]{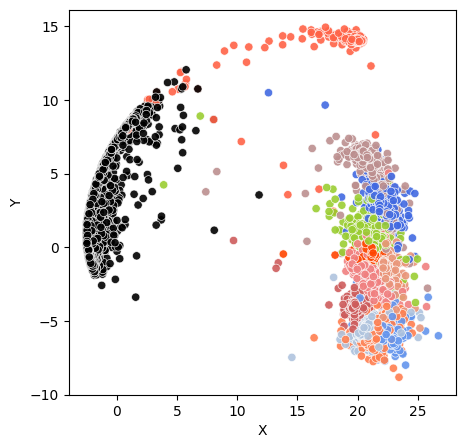}}%
\hfill
\subfloat[RoBERTa-Mixset.]{\label{fig:rob_mix}  
\includegraphics[width=0.2355\textwidth]{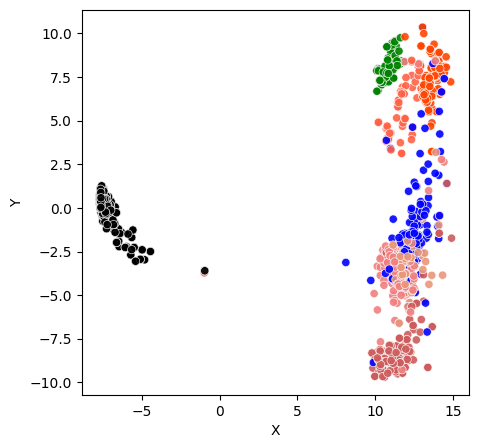}}%
\hfill
\subfloat[{\sc \textbf{TopFormer}}-Mixset.]{\label{fig:toprob_mix}  
\includegraphics[width=0.23\textwidth]{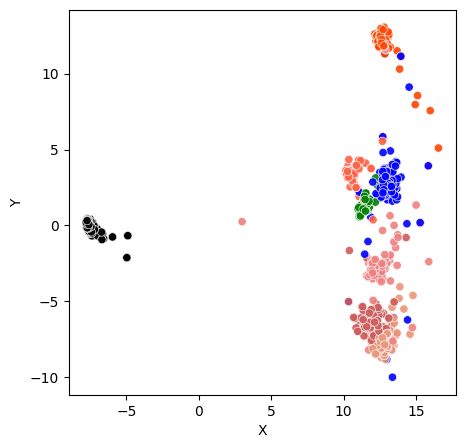}}%

\caption{PCA plots from RoBERTa and {\sc \textbf{TopFormer}} training embeddings for the SynSciPass and Mixset datasets on the classification tasks. 
    The \textbf{black} clusters are the human labels and the other clusters are the deepfake text labels.}
    \label{fig:pca}
    \vspace{10pt}
\end{figure*}

\section{Discussion} \label{discussion}
% See below for the observed strengths and weaknesses of adding a Topological layer
% to a Transformer-based model:
%\begin{enumerate}[leftmargin=\dimexpr\parindent-0.2\labelwidth\relax,noitemsep]

    \subsection{Improvement with TDA is Not Random}
    Since RoBERTa is a black-box model, to confirm that {\sc TopFormer}'s performance is not due to training with the right kind of ``noise,''
    we compare with the Gaussian model - Gaussian-RoBERTa. 
    The hypothesis is that if TDA's performance is due to noise, then the Gaussian models trained on noise should perform similarly. 
    We observe from  Tables \ref{tab:llm_results}, \ref{tab:syn_results}, and \ref{tab:m_results} that
    Gaussian-RoBERTa underperforms for both OpenLLMText and Mixset and marginally improves by 1\% for the SynSciPass dataset.
    While Gaussian-BERT underperforms for all 3 datasets in Table \ref{tab:bert}. 
    However, {\sc TopFormer} w/ BERT outperforms their base models for 2 out of 3 datasets, improving by a larger margin on the Mixset dataset (5\%, Table \ref{tab:bert}).
    In addition, we observe that the baseline SOTA deepfake AA models - GPT-who and Contra-BERT underperform in the task. 
    Finally, {\sc TopFormer} is the only model that was able to consistently outperform all other models. 
    Thus, the large decrease in performance in the Gaussian models for the 3 datasets, suggests that our TDA-based techniques performances are not by chance.

    \subsection{TDA Captures Structural Features
    which Complement RoBERTa Weights}
    There are currently 5 linguistic levels - 
    phonology, pragmatics, morphology, syntax, and 
    semantics. Phonology has to do with spoken speech 
    sounds and pragmatics is how language is used to 
    convey meaning. The most relevant for written text 
    classifications are syntax, morphology, and semantics. This means 
    that to accurately capture the linguistic features 
    from a piece of text, syntactic, morphology, and 
    semantic features are needed. 
    However, RoBERTa captures a broad range of linguistic
    patterns, such as contextual representations in terms of syntactic and semantic relationships. 
    This means that on the surface level, RoBERTa is only able to capture word order and thus can infer syntactic and some semantic relationships. However, 
    TDA features capture the true shape or structure of data even with the presence of noise. These features in the context of NLP, can be interpreted as linguistic structures. Finally, by combing these 3 linguistic features, our model - {\sc TopFormer} can more accurately distinguish deepfake texts 
    from human-written ones, as observed in Tables \ref{tab:llm_results},  \ref{tab:syn_results}, \ref{tab:m_results} and \ref{tab:style}. 

    This can be further observed in Figures \ref{fig:pca}, 
    where we plot the PCA features which is a linear 
    combination of RoBERTa's word embeddings vs. {\sc TopFormer}'s word embeddings of the SynSciPass and Mixset datasets (i.e., the 2 datasets where {\sc TopFormer} outperformed RoBERTa, the backbone by a high margin).  
    We observe the more distinct clusters in {\sc TopFormer}'s plots for both datasets. This suggests that TDA can extract additional features for more accurate attribution of authors. 

\renewcommand{\tabcolsep}{0.4pt}
\begin{table}[tb]
    \centering
    \footnotesize
    \begin{tabular}{ccccccc}
      \toprule
       \multirow{3}{*}{\textbf{MODEL}} & \multicolumn{2}{c}{\textbf{OpenLLMText}} 
       & \multicolumn{2}{c}{\textbf{SynSciPass}} & 
       \multicolumn{2}{c}{\textbf{Mixset}} \\
      \cmidrule(lr){2-3} \cmidrule(lr){4-5} \cmidrule(lr){6-7}
      &  
      \textbf{Macro F1} & \textbf{\%$\Delta$} &
      \textbf{Macro F1} & \textbf{\%$\Delta$} & 
      \textbf{Macro F1} & \textbf{\%$\Delta$} \\
      \cmidrule(lr){2-7}
      RoBERTa & 0.7288 & - & 0.8719 & - & \underline{0.7705}  & - \\
      
      \textsc{\textbf{TopFormer}}\_{attn} &  \textbf{0.8970} & \textcolor{blue}{17\% $\uparrow$} & \underline{0.8923}  & \textcolor{blue}{2\% $\uparrow$} 
      &  0.6438 & \textcolor{red}{8\% $\downarrow$}  \\
      
      \textsc{\textbf{TopFormer}}\_{hidden} & 0.1961 & \textcolor{red}{53\% $\downarrow$} 
      & 0.0794 & \textcolor{red}{ 79\% $\downarrow$}
      & 0.0955 & \textcolor{red}{72\% $\downarrow$} \\
      
      \textsc{\textbf{TopFormer}}\_{pool\_corr} & \underline{0.7993} & \textcolor{blue}{7\% $\uparrow$} 
      & 0.0794 & \textcolor{red}{79\% $\downarrow$}
      & 0.2045 & \textcolor{red}{53\% $\downarrow$} \\
      
      % {\textsc{TopFormer}\_{pool} (\textbf{Ours}) & 0.7522 & \textcolor{blue}{2\% $\uparrow$}  
      % & \textbf{0.9058} & \textcolor{blue}{4\% $\uparrow$} & 
      % \textbf{0.8294} & \textcolor{blue}{6\% $\uparrow$} \\
      \cmidrule(lr){1-7}
      \textsc{\textbf{TopFormer}} (\textbf{Ours}) & 0.7522 & \textcolor{blue}{2\% $\uparrow$}  
      & \textbf{0.9058} & \textcolor{blue}{4\% $\uparrow$} & 
      \textbf{0.8294} & \textcolor{blue}{6\% $\uparrow$} \\

      \bottomrule 
    \end{tabular}
    \caption{Ablation study results. The best performance is \textbf{boldened} and the second best is \underline{underlined}. \%$\Delta$ is the \% Gains in Macro F1.}
    \label{tab:ablation}
    \vspace{10pt}
\end{table}

    \subsection{{\sc TopFormer} Performs Well on Style Detection}
    To evaluate the robustness of 
    {\sc TopFormer} to data with multi-style labels, 
    we run further experiments to compare vanilla RoBERTa to 
    {\sc TopFormer}. 
    For this task, we use the style labels for each dataset. 
    OpenLLMText and Mixset have the same 3 labels 
    - \textit{human, paraphrasers,}
    and \textit{generators}, while SynSciPass has an additional label to the 3 labels - \textit{translators}. See Table \ref{tab:style} for results.
    % We observe that {\sc TopFormer} w/ BERT achieves high performance in this task for the OpenLLMText and Mixset datasets (2\% and 4\% increases, respectively). 
    For the SynSciPass dataset, we observe that {\sc TopFormer} 
    significantly outperforms the baseline (7\% increase, Table \ref{tab:style}). 
    This could be because the SynSciPass dataset has an additional writing style (translators) than the other datasets, and since TDA performs well when data is noisy and heterogeneous, this higher variability improves performance. 
    % We also observed that where TopBERT or TopRoBERTa does outperform, it performs comparably, suggesting its utility in style detection. 
    % Overall, there is still no loss in performance when TDA-based techniques do not outperform. 

    \subsection{{\sc TopFormer} Performs comparably on Homogeneous Datasets}
    To further evaluate where {\sc TopFormer} performs well and underperforms, 
    we run further experiments on homogeneous datasets - TuringBench (TB) and M4, which have only 2 distinct writing styles (i.e., human and deepfake generation). 
    %TDA thrives in high variability, we investigated the scenario where it might not thrive - low variability. 
    We observe from Table \ref{tab:error} that {\sc TopFormer} underperforms and outperforms marginally by 1\% for both datasets, 
    always outperforming Gaussian-RoBERTa. 
    This suggests that even when {\sc TopFormer} does not perform optimally, it still performs comparably to RoBERTa, suggesting no huge loss in performance.

    \subsection{Larger is Not Always Better: Reshaped $pooled\_output$ is a Better TDA Input}
    Most researchers that apply TDA for NLP tasks commonly use word2vec embeddings or attention 
    weights as input to extract TDA features \cite{perez2022topological,kushnareva2021artificial,wu2022topological,haghighatkhah2022story,doshi2018movie,savle2019topological}. 
    We observe that 
    while using the attention weights, the last hidden state, and the correlation matrix of the $pooled\_output$ as input is more intuitive as they are already 
    in the right format for the TDA layer ($>$ 1D vector), 
    the reshaped 2D matrix of the $pooled\_output$ contains richer features for the TDA layer. See Table \ref{tab:ablation} for the ablation results of different inputs for the TDA layer. 
    In confirmation with \citet{kushnareva2021artificial}, we discover that using all the inputs except for the reshaped $pooled\_output$ to extract TDA features is unstable. 
    This could be why previous studies first constructed directed and undirected graphs with the attention weights before TDA feature extraction to encourage stability \cite{perez2022topological,kushnareva2021artificial}. 
    Nevertheless, these techniques significantly increase 
    computational costs. 
    
    The attention weights are large matrices, such as 
    $Max\_length \times 768$, yielding a
    $1 \times 1533$ TDA output, which increases the latent space of the Linear layer 
    from 768 to 2301. 
    While the last hidden state and the correlation matrix increase the latent spaces to 19164 and 3068, respectively. 
    Additionally, to maintain consistent dimensions for TDA 
    features, we employ normalization techniques when TDA features for some articles 
    differ from the majority. In contrast, our technique increases the dimensions 
    of the linear layer from 768 to 837 for all datasets without requiring normalization. 
    Finally, the results of the ablation study in 
    Table \ref{tab:ablation} suggests that our reshaped $pooled\_output$ technique is superior as {\sc TopFormer}
    consistently outperformed in all 3 datasets. Even though, 
    we observe a surprising 
    increase in performance for {\sc TopFormer}\_{attn} and {\sc TopFormer}\_{pool\_corr} 
    evaluated on the OpenLLMText dataset. This still confirms the utility of TDA, while suggesting that \textit{\sc TopFormer (Ours)} is the most consistent technique and computationally less expensive than other techniques.

\section{Conclusion and Future Work}
We propose a novel solution to accurately 
attribute the authorship of deepfake vs. human texts - {\sc \textbf{TopFormer}}. 
This technique entails including a Topological layer in a Transformer-based model, such that the Linear layer's input is a concatenation of the backbone's regularized weights and the TDA features. 
Next, we evaluate our model on realistic datasets that reflect the landscape of how people use LLMs, either to generate, paraphrase, summarize, or edit a piece of texts. 
%However, initial benchmark datasets typically focus on the generation technique. Furthermore, these datasets do not reflect the gross label imbalance with humans authoring most of the text in the information space; the datasets typically have a uniform representation of labels or sometimes even more deepfake labels than human labels, falsely representing the current landscape. 
Our novel technique, {\sc {TopFormer}} outperforms all SOTA baseline models
when evaluated on these 3 realistic datasets. 
% preparing for the current landscape which will be further exacerbated in the future. 

Lastly, in the future, we will scrutinize our models under stricter constraints such as evaluation on 
adversarial robustness, known as \textit{Authorship Obfuscation}, open-set datasets, and out-of-distribution datasets, such as low-resource languages, 
multilingual, and insufficiently sized datasets.

\section{Ethical Statement}
Since the advent of LLMs, society has had to grapple with the many benefits and huge potential malicious uses of such technology. 
To mitigate these risks, some being obvious security risks like the creation of authentic-looking misinformation, researchers have been working on several niche areas in LLM, including \textit{deepfake text detection}. However, due to the opposing benefits of LLMs (i.e., great good and great evil), we understand that the proposed mitigation of risks that LLMs pose can also be turned around and used for evil. 
Thus, we understand that detection models like {\sc {TopFormer}} which can more accurately detect deepfake texts, especially in stricter conditions (i.e., imbalanced sample, multi-style labels, noisy data), can also be used to test when deepfake texts can evade detection after obfuscation. However, due to the serious security risks that LLMs pose, we believe that the benefits of accurate authorship attribution of deepfake text authors outweigh the potential risks.

\section*{Acknowledgement}
This work was in part supported by NSF awards \#1820609, \#2114824, and \#2131144.
Adaku thanks Dr. Bastian Rieck 
for the insightful discussions on how to improve our TDA technique. 
Also, Adaku would like to thank Dr. Charlie Dagli for
reading the paper drafts and providing invaluable recommendations.

%%%%%%%%%%%%%%%%%%%%%%%%%%%%%%%%%%%%%%%%%%%%%%%%%%%%%%%%%%%%%%%%%%%%%%%%

%%% Use this environment to include acknowledgements (optional).
%%% This will be omitted in doubleblind mode.

% \begin{ack}
% This work was in part supported by NSF awards \#1820609, \#2114824, and \#2131144.
% Adaku thanks Dr. Bastian Rieck 
% for the insightful discussions on how to improve our TDA technique. 
% Also, Adaku would like to thank Dr. Cagri (Charlie) Dagli for
% reading this paper's drafts and providing invaluable recommendations. 
% \end{ack}

%%%%%%%%%%%%%%%%%%%%%%%%%%%%%%%%%%%%%%%%%%%%%%%%%%%%%%%%%%%%%%%%%%%%%%%%

% \newpage
%%% Use this command to include your bibliography file.

\bibliography{ecai}

\end{document}